\documentclass[runningheads]{llncs}
\usepackage{graphicx}
\usepackage[T1]{fontenc}
\usepackage{verbatimbox,lipsum}
\usepackage{subcaption}
\usepackage[utf8]{inputenc}
\usepackage{listings}
\usepackage{pifont}
\usepackage{float}
\usepackage{tablefootnote}
\newfloat{lstfloat}{htbp}{lop}
\floatname{lstfloat}{Listing}

\usepackage[most]{tcolorbox}
\usepackage{array}
\usepackage{amsmath}
\usepackage{xcolor}
\usepackage{tikz}
\usepackage{colortbl}
\usepackage{multirow}
\usepackage{booktabs}
\usepackage{amssymb}

\begin{document}

\title{Nested Named Entity Recognition in Plasma Physics Research Articles}

\author{Muhammad Haris\inst{1}\orcidID{0000-0002-5071-1658} \and
Hans Höft\inst{2}\orcidID{0000-0002-9224-4103} \and
Markus M. Becker\inst{2}\orcidID{0000-0001-9324-3236} \and
Markus Stocker\inst{1}\orcidID{0000-0001-5492-3212}
}

\authorrunning{Haris et al.}

\institute{TIB---Leibniz Information Centre for Science and Technology, Leibniz University Hannover, 30167, Hannover, Germany\\
\email{\{muhammad.haris, markus.stocker\}@tib.eu} \and
Leibniz Institute for Plasma Science and Technology (INP), Felix-Hausdorff-Str. 2, 17489 Greifswald, Germany\\
\email{\{markus.becker, hans.hoeft\}@inp-greifswald.de}}

\maketitle              

\begin{abstract}
Named Entity Recognition (NER) is an important task in natural language processing that aims to identify and extract key entities from unstructured text. We present a novel application of NER in plasma physics research articles and address the challenges of extracting specialized entities from scientific text in this domain. Research articles in plasma physics often contain highly complex and context-rich content that must be extracted to enable, e.g., advanced search. We propose a lightweight approach based on encoder-transformers and conditional random fields to extract (nested) named entities from plasma physics research articles. First, we annotate a plasma physics corpus with 16 classes specifically designed for the nested NER task. Second, we evaluate an entity-specific model specialization approach, where independent BERT–CRF models are trained to recognize individual entity types in plasma physics text. Third, we integrate an optimization process to systematically fine-tune hyperparameters and enhance model performance. Our work contributes to the advancement of entity recognition in plasma physics and also provides a foundation to support researchers in navigating and analyzing scientific literature.
\keywords{Nested Named Entity Recognition \and Scholarly Communication \and Domain-specific NER \and Information Extraction \and Bayesian Optimization \and Hyperparameters tuning}
\end{abstract}

\section{Introduction}
Named Entity Recognition (NER), a branch of natural language processing (NLP), provides a promising solution to automatically identify and classify named entities in unstructured text into predefined categories such as locations, organization, persons etc~\cite{tjong-2003-introduction}. This technology is widely adopted in several academic fields including, computer science~\cite{csner}, life sciences~\cite{biomedner}, agriculture~\cite{d2024agriculture}, and biodiversity~\cite{biodivner}. 

Plasma physics, a research field that studies the physical properties of plasmas, reports scientific findings about complex phenomena with rich contextual information including experimental details. The specialized, complex and context-rich terminology used in plasma physics research articles makes it difficult for systems to efficiently support researchers in searching and analyzing relevant information. NER offers a solution to automatically identify and classify terminology and thus support researchers with more efficient information retrieval and knowledge discovery. In this work, we employ Nested NER (NNER), because in addition to being complex and context-rich, terminology used in plasma physics often carries multiple meanings or belongs to overlapping categories, two challenges that NNER can address.

Several frameworks have been developed to extract nested entities from natural language text, including span-based  methods~\cite{yan-etal-2023-embarrassingly,shibuya2020nested}, hypergraph models~\cite{lu2015joint,katiyar2018nested,Yan_Cai_Song_2023}, and layered approaches~\cite{ju-etal-2018-neural,wang-etal-2020-pyramid,rehana2024nested}. While these techniques show potential, applying NER to plasma physics texts remains a significant challenge. The diversity of entity classes, the complexity of domain-specific terminology, and uneven data distribution requires a specialized approach. Most existing NER systems are designed for general or biomedical text and they struggle to accurately capture the important entities/mentions from plasma physics literature, specifically. Even with the advent of advanced models, such as layered approaches and hypergraph models, satisfactory performance on domain-specific datasets cannot be guaranteed. Underrepresented classes are often overlooked, which leads to less stable prediction performance. Given that manual data annotation is both time-consuming and labor-intensive, there is a pressing need for methods that can effectively handle imbalanced data while maintaining high accuracy, as this is essential for ensuring the usability of extracted information in real-world applications.

This paper addresses these challenges by developing a NNER model specifically designed for plasma physics. Our contributions are summarized as follows:
\begin{enumerate}
    \item We annotate and publish a domain-specific plasma physics dataset comprising 16 entity classes created for Named Entity Recognition (NER) tasks.
    \item We address the problem of NNER in the plasma physics domain by proposing a lightweight BERT-CRF-based model. Our approach employs entity-specific specialization, in which independent BERT–CRF models are trained for individual entity types and their predictions are subsequently aggregated to produce the final outputs. Fine-tuning a pretrained language model (e.g., BERT) with this architecture allows the system to learn class-specific patterns while leveraging shared knowledge across all entity types.
    \item To further enhance model performance, we incorporate Bayesian Optimization (BO) to automatically fine-tune the hyperparameters of our NER model, which leads to improved evaluation metrics.
    \item We evaluate the proposed approach and also experiment with several baseline models on our annotated dataset to demonstrate the reliability of our approach. To further assess the generalizability of our method, we also evaluate it on publicly available datasets, namely GENIA and the Chilean Waiting List corpus.
\end{enumerate}

\section{Related Work}
Ju et al.~\cite{ju-etal-2018-neural} have introduced a dynamic neural model for NNER that operates without the need for external knowledge resources or handcrafted linguistic features. This approach enables the sequential stacking of flat NER layers from bottom to top, which allows for end-to-end identification of nested entities.

Pyramid et al.~\cite{wang-etal-2020-pyramid}, a layered model designed for NNER, is another notable approach. This architecture is composed of a stack of interconnected layers, where each layer is responsible for predicting whether a text span corresponds to a complete entity mention.

Fu et al.~\cite{Fu_Tan_2021} frame NNER as a constituency parsing task with partially observed trees that leverages a partially observed TreeCRF to model this structure. In this approach, all labeled entity spans are treated as observed nodes within a constituency tree, while the remaining spans are considered latent. The use of TreeCRFs provides a unified framework for jointly modeling both the observed and latent nodes, enabling effective handling of the hierarchical structure inherent in nested entities.

Rehana et al.~\cite{rehana2024nested} present a BERT-based approach for NNER, specifically leveraging the pretrained PubMedBERT model to address the challenges posed by nested entities in biomedical texts. The method integrates rich contextual embeddings from PubMedBERT with a multilayer tagging strategy, enabling the model to effectively distinguish overlapping entity mentions.

To address the challenge of limited overlap between the training and test sets, the approach by Wan et al.~\cite{wan-etal-2022-nested} focuses on enhancing span representations through retrieval-based span-level graphs. This method constructs global entity-entity and span-entity graphs by leveraging n-gram similarity, effectively linking spans in the input with semantically similar entities from the training data.

A novel sequence-to-set neural framework~\cite{tan21sequence} has also been proposed for NNER, an approach that aims to move beyond predefined candidate spans. This approach introduces a fixed set of learnable vectors that are trained to capture patterns associated with meaningful spans. By employing a non-autoregressive decoder, the model predicts the entire set of entities in a single forward pass.

Yang et al.~\cite{yang-etal-2022-nested} proposed an NNER approach that models all nested entities within a sentence as a unified, holistic structure. The authors proposed a holistic structure parsing algorithm to extract the complete set of entities in a single pass.

A study by Rojas et al.~\cite{rojas-etal-2022-simple} revisits LSTM-CRF and proposed Multiple LSTM-CRF (MLC) model, a straightforward yet under-explored approach that trains separate sequence labeling models independently for each entity type. The extensive experiments conducted on three NNER datasets demonstrate that MLC achieves competitive performance.

Finally, Lu and Roth~\cite{lu2015joint} introduced a hypergraph representation for extracting entity mentions. Katiyar and Cardie~\cite{katiyar2018nested} proposed a directed hypergraph model leveraging LSTM-based features to learn nesting patterns. Muis and Lu~\cite{muis2017labeling} proposed another approach by incorporating mention separators and handcrafted features.

While prior work has primarily focused on NER methods for domain-agnostic datasets and well-studied domains such as biomedical corpora, domain-specific challenges and the role of automatic hyperparameter tuning have received comparatively less attention. In contrast, our work introduces a curated dataset for the plasma physics domain, develops a specialized NER approach designed for this domain, and integrates Bayesian Optimization to automatically tune hyperparameters. This combination distinguishes our contribution from existing approaches, which typically either emphasize generic modeling strategies or rely on manually selected hyperparameters.

\section{Methodology}
\subsection{Problem Formulation}
We define the Named Entity Recognition (NER) task as a token (words) classification problem. Given an input sentence \( x \in \mathcal{X} \), where $ \mathcal{X}$ is the set of all possible token sequences and \( x = \{w_1, w_2, \dots, w_n\} \) consists of \( n \) tokens, the objective is to assign a label \( y_i \) from a predefined set of entity labels \( \mathcal{E} \) to each token \( w_i \). 

Each token \( w_i \) is assigned a label \( y_i \) such that \( y_i \in \mathcal{E} = \{O, B_{E_1}, I_{E_1}, B_{E_2}, I_{E_2}, \dots, B_{E_m}, I_{E_m} \} \), where \( O \) represents tokens outside any entity, and \( B_{E_k}, I_{E_k} \) with \(k = 1, \dots, m\) denote the beginning and inside tags for entity type \( E_k \). Since a token may belong to multiple overlapping entity types, we model this as a multi-label classification problem, where each token can be associated with a subset of entity labels, i.e., \( y_i \subseteq \mathcal{E} \) with \( |y_i| \geq 1 \) in cases of overlapping entities.
\subsection{Dataset Preparation}
Our corpus consists of plasma physics research articles and patents. Domain experts annotated the dataset using the INCEpTION tool\footnote{\url{https://inception-project.github.io/}} with 16 predefined entity classes specifically designed for Named Entity Recognition (NER) tasks. In total, the corpus includes 30 full-text research papers and 500 patent abstracts, comprising 10,272 sentences. Document selection was not arbitrary, we have followed domain-specific heuristics to ensure diversity in experimental configurations (e.g., dielectric barrier discharge), target applications, and device types. Entity types were defined based on well-established concepts from the Plasma-MDS metadata schema~\cite{franke2020plasma}, that ensures semantic consistency. The selected articles are representative of a particular subfield in low-temperature plasma physics. The defined entity types capture overarching domain concepts, meaning that the annotation data can be considered broadly representative of non-thermal gas discharges at atmospheric pressure (e.g.~\cite{Kogelschatz-2003-ID2019,Kettlitz-2012-ID5914}), even though certain subfields such as plasma medicine are not yet covered in detail. In addition, the scope of the patent corpus is broader and not limited to atmospheric-pressure discharges.
\begin{figure*}[t!]
  \centering
  \includegraphics[width=0.70\textwidth]{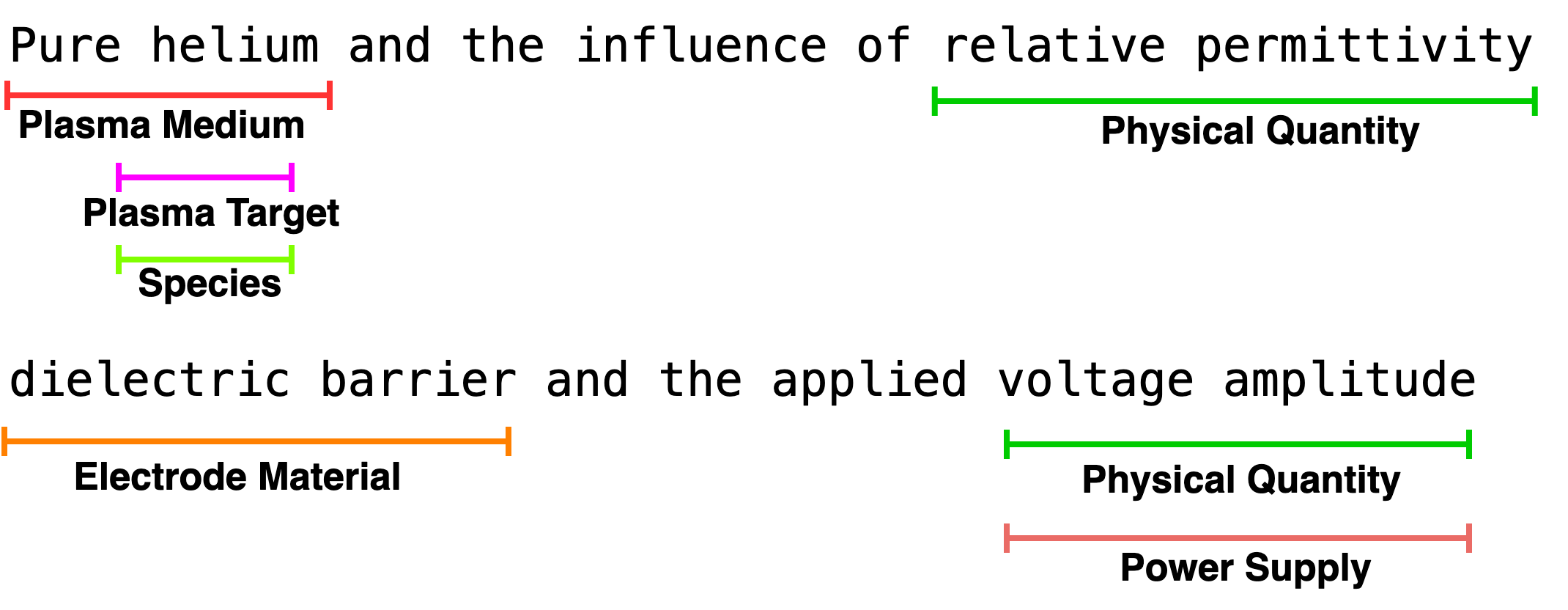}
  \caption{An example of nested entities from the annotated dataset, where the text is labeled with multiple overlapping entity types for specific terms.}
  \label{fig1}
\end{figure*}

\begin{table}[t!]
\setlength{\tabcolsep}{5pt}
\centering
\caption{Entity types used in the dataset, with concise definitions and representative examples.}
\small
\begin{tabular}{p{3cm} p{4cm} p{4cm}}
\toprule
\textbf{Entity Type} & \textbf{Definition} & \textbf{Examples} \\
\midrule
Physical Quantity &
Observable property of a physical object, event or system&
mobility, velocity, electron energy \\

Physical Effect &
Phenomenon leading to a measurable change in a physical system governed by the laws of physics&
diffusion, ionization, heating \\

Species &
Charged or neutral particles with a medium &
ozone, electron, molecular nitrogen \\

Diagnostic Device &
Instrument used to observe physical quantities &
ICCD camera, fiber-optic spectrometer, oscilloscope \\

Plasma Source &
Device or configuration used to generate plasma &
dielectric barrier discharge, plasma jet, spark discharge\\

Power Supply &
System or component that provides electrical power &
sinusoidal operation, pulsed operation, voltage amplitude \\

Plasma Medium &
Medium or environment in which plasma is generated &
air, feeding gas mixture, helium \\

Experiment &
Specific physical setup for scientific measurements &
optical emission spectroscopy, iCCD recording, electrical measurement\\

Electrode Material &
Medium acting as an electrode&
tungsten, stainless steel, alumina \\

Plasma Application &
An application domain where plasma is used. &
ozone synthesis, sterilization, plasma display \\

Unit &
Standardized quantity used to express measurements of physical quantities &
kV, kHz, mm \\

Modelling &
Theoretical, numerical, or computational study &
fluid model, reaction kinetics model, particle-in-cell simulation\\

Discharge Regime &
Operating mode of an electrical discharge &
diffuse, filamentary discharge, glow discharge \\

Electrode Configuration &
Geometric or structural electrode arrangement &
single filament, embedded mesh, coaxial\\

Plasma Properties &
Intrinsic properties describing plasma state &
atmospheric pressure, non-thermal, partly ionized\\

Plasma Target &
Material or object exposed to plasma treatment &
liquid surface, bacterial suspension, wafer \\
\bottomrule
\end{tabular}
\label{table1}
\end{table}

Nested annotations play a crucial role in our corpus. An entity is considered nested if its span overlaps with another labeled entity in a different annotation layer. When an entity is embedded within another entity, both are annotated with their respective labels. Figure~\ref{fig1} illustrates an example of nested named entities, where the term \textit{pure helium} refers to a \texttt{Plasma Medium} entity that overlaps with another term, which is labeled as \texttt{Plasma Target}. Similarly, the term \textit{voltage amplitude} refers to different classes that is why it is annotated with both \texttt{Physical Quantity} and \texttt{Power Supply} classes. Table~\ref{table1} presents entity class definitions with examples, while Table~\ref{table2} reports detailed descriptions and frequency statistics for each class. We report the number of sentences, total entities, and nested entities for each data split. The nested percentage indicates the proportion of entities involved in overlapping mentions.

In data preparation, we formatted the dataset using the BIO (Begin-Inside-Outside) tagging scheme for each entity type. For every type, we filtered the training data to retain only relevant annotations and labeling all unrelated tokens as \textit{O} (non-entity). We divided the dataset of each type into three different datasets: training (70\%), validation (15\%) and test (15\%).
\begin{table}[t!]
\centering
\small
\setlength{\tabcolsep}{4pt}
\renewcommand{\arraystretch}{1.05}
\caption{Statistics of the dataset: showing frequency of each entity as well as number of nested entities.}
\begin{tabular*}{\textwidth}{@{\extracolsep{\fill}} l r r r r r r @{}}
\hline
\textbf{Item} & \textbf{Total} & \multicolumn{2}{c}{\textbf{Nested Entities}} & \textbf{Train} & \textbf{Test} & \textbf{Val} \\
\cline{3-4}
 &  & \textbf{Count} & \textbf{\%} &  &  &  \\
\hline
Sentences        & 10,272 & --   & --     & 7,191 & 1,542 & 1,539 \\
Split percentage & --     & --   & --     & 70\%  & 15\%  & 15\%  \\
\hline
Physical Quantity & 19710 & 4706 & 23.87\% & 13669 & 3121 & 2920 \\
Physical Effect & 11164 & 3036 & 27.19\% & 7799 & 1738 & 1627 \\
Species & 8935 & 2196 & 24.57\% & 6069 & 1530 & 1336 \\
Diagnostic Device & 8348 & 1447 & 17.33\% & 6067 & 1211 & 1070 \\
Plasma Source & 6412 & 1984 & 30.94\% & 4570 & 964 & 878 \\
Power Supply & 5463 & 1275 & 23.33\% & 3825 & 867 & 771 \\
Plasma Medium & 4857 & 1041 & 21.43\% & 3275 & 827 & 755 \\
Experiment & 4528 & 1211 & 26.74\% & 3198 & 688 & 642 \\
Electrode Material & 4149 & 1091 & 26.29\% & 2836 & 706 & 607 \\
Plasma Application & 4131 & 836 & 20.23\% & 2803 & 691 & 637 \\
Unit & 2929 & 1563 & 53.36\% & 2078 & 442 & 409 \\
Modelling & 2665 & 605 & 22.70\% & 1891 & 408 & 366 \\
Discharge Regime & 2643 & 970 & 36.70\% & 1872 & 399 & 372 \\
Electrode Configuration & 2576 & 718 & 27.87\% & 1849 & 336 & 391 \\
Plasma Properties & 2338 & 858 & 36.69\% & 1599 & 378 & 361 \\
Plasma Target & 856 & 323 & 37.73\% & 594 & 125 & 137 \\
\hline
\end{tabular*}
\label{table2}
\end{table}
\subsection{NNER Model Construction}
\begin{figure*}[t!]
  \centering
  \includegraphics[width=\textwidth]{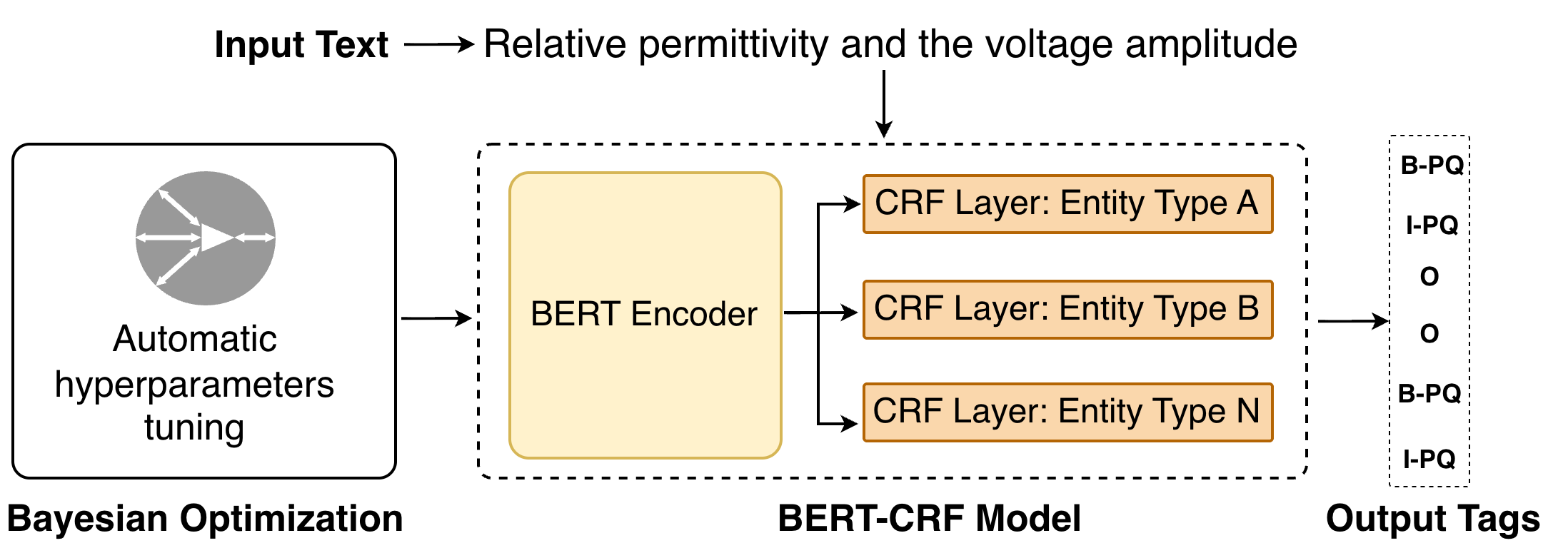}
  \caption{Overview of the proposed BERT–CRF approach with Bayesian Optimization (BO). BO operates in an outer loop to tune hyperparameters, while specialized BERT–CRF models are trained for individual entity types and their predictions are aggregated to handle nested entities.}
  \label{fig2}
\end{figure*}

For NNER, we adopt a BERT-CRF architecture (Figure~\ref{fig2}).Bidirectional Encoder Representations from Transformers (BERT)~\cite{devlin-etal-2019-bert} is well known for producing deep contextualized word embeddings that significantly improve natural language understanding tasks. However, while BERT captures contextual semantics, it does not explicitly model structural dependencies between output labels, which can lead to invalid or inconsistent entity sequences. To address this limitation, we add a Conditional Random Field (CRF) layer~\cite{Lafferty-crf} on top of BERT. The CRF layer models dependencies between neighboring labels that reduce inconsistencies in the output sequence and provide more consistent predictions, which is particularly beneficial for structured NER tasks involving nested and overlapping entities.

Our model is inspired by Multi-head Dense-augmented CRF~\cite{multi-head-crf} and Multiple LSTM-CRF (MLC)~\cite{rojas-etal-2022-simple} approaches. However, instead of relying on additional middleware, we employ entity-specific BERT–CRF models built directly on pretrained transformer representations. We further integrate Bayesian Optimization (BO) for tuning key hyperparameters which improves performance under imbalanced and nested entity settings.

Our dataset exhibits significant class imbalance, with entity frequencies ranging from 856 to 19,710 instances. In the presence of overlapping and nested entities, training a single-head sequence labeling model across all entity types can lead to performance degradation due to entity-type interference, particularly for rare classes or for classes that represent shorter sequences. Rather than collecting or annotating additional data (which is time-consuming), we adopt a more moderate approach: leveraging a BERT–CRF framework that supports per-entity-type specialization while benefiting from shared contextual representations.

In the presence of overlapping and nested entities, training a single sequence labeling model across all entity types can lead to entity-type interference, particularly for rare classes or for entities represented by short spans. Rather than collecting or annotating additional data, which is time-consuming, we adopt a more moderate approach: leveraging a BERT–CRF framework with per-entity-type specialization. This design enables the model to better accommodate class imbalance and span variability and supports stable and competitive performance under strict span-based evaluation.

This design effectively reduces label conflicts by decoupling decision layers. Each CRF head focuses on its respective entity distribution, that improves learning for underrepresented classes. Such a method also improves the interpretability of model behavior and performance for each entity type, as we can analyze the errors, predictions, and learning aspects of each model.

\subsection{Automatic Fine-Tuning Model Parameters}
Given a BERT-CRF model for NNER, we aim to optimize hyperparameters 
\(\theta = (\text{learning\_rate}, \text{batch\_size}, \text{weight\_decay})\) to maximize the F1-score on the validation set:

\[
\theta^* = \arg\max_{\theta} F_1(\theta),
\]

where \(F_1(\theta)\) is the F1-score obtained by training the model with hyperparameters \(\theta\). To achieve optimal hyperparameters, we choose Bayesian Optimization (BO)~\cite{jones1998efficient}, a method for optimizing black-box functions that are costly to evaluate. It is suitable in high-dimensional spaces and leverages Gaussian Processes (GPs)~\cite{rasmussen2003gaussian} to model the objective function and an acquisition function to guide the search.

We employ Gaussian Process (GP) surrogate modeling to efficiently optimize hyperparameters \(\theta\) by approximating the relationship between configurations and their corresponding validation F1-scores \(F_1(\theta)\).  The GP provides a probabilistic representation of \(F_1(\theta)\) through a mean function \(\mu(\theta)\) and a covariance kernel \(k(\theta, \theta')\), that enables simultaneous estimation of predicted performance and uncertainty across the hyperparameter space.

As a sequential optimization strategy for expensive-to-evaluate functions, BO leverages this GP surrogate trained on observed data 
\[
\mathcal{D} = \{ (\theta_i, F_1(\theta_i)) \}_{i=1}^m
\]
to infer the posterior distribution over the objective function, \(P(F_1 \mid \mathcal{D})\). 
The optimization process is guided by an acquisition function \(\alpha(\theta)\) that quantifies the utility of evaluating each candidate configuration 
\(\theta \in \Theta\), where \(\Theta\) denotes the feasible hyperparameter space. 
This two-stage approach ensures computational efficiency, as the acquisition function operates on the surrogate model rather than directly evaluating the expensive objective function. To guide the hyperparameter search, we maximize the \textit{Expected Improvement (EI)} over the current best F1-score (\(F_1^+\)):

\[
\text{EI}(\theta) = \mathbb{E}\left[\max(F_1(\theta) - F_1^+, 0)\right]
\]

where \(F_1(\theta)\) is the predicted F1-score for hyperparameters \(\theta\) and \(F_1^+\) is the highest observed F1-score so far. This criterion balances exploration (sampling high-uncertainty regions where \(F_1(\theta)\) may exceed \(F_1^+\)). Under the Gaussian Process surrogate, EI has a closed-form solution that combines the predicted mean \(\mu(\theta)\) and uncertainty \(\sigma(\theta)\) of \(F_1(\theta)\).

We leverage this approach to automatically fine-tune the hyperparameters of our model and, thus, optimize performance in computationally intensive tasks such as training deep learning models. By modeling uncertainty and focusing the evaluation on high-utility regions, Bayesian Optimization offers a data-efficient and scalable alternative to traditional methods such as grid or random search, and can be accelerated with GPU support for large-scale experiments.

\section{Experiments}
\label{s:experiments}
We evaluate the effectiveness and generalizability of our model on the annotated plasma physics dataset and compare it against several state-of-the-art NER methods. We further assess generalization on two widely used benchmarks: GENIA~\cite{kim2003genia} and the Chilean Waiting List~\cite{baez2020chilean}.

\subsection{Evaluation on the Plasma Physics Dataset}
We evaluate our approach on the annotated plasma physics dataset introduced in this work and compare performance with established NNER methods under a consistent experimental setup.

\begin{itemize}
    \item Span-based and biaffine~\cite{yan-etal-2023-embarrassingly}: Utilizes a biaffine model to score pairs of start and end tokens within a sequence for entity span detection.

    \item Pyramid~\cite{wang-etal-2020-pyramid}: Introduces normal and inverse pyramidal structures to capture bidirectional interactions for entity recognition.
    
    \item Partially-Observed Tree CRFs~\cite{Fu_Tan_2021}: Employs a partially observed TreeCRF to jointly model these nodes, that effectively capturing hierarchical nested entity structures.

    \item Layered model~\cite{ju-etal-2018-neural}: A sequence labeling-based model designed to identify nested entities by dynamically stacking LSTM-CRF layers.

    \item Second-best~\cite{shibuya2020nested}: Applies second-best sequence decoding to detect nested entities within already extracted spans.

    \item Span-based Attention Network~\cite{yuan-etal-2022-fusing}: Triaffine attention encodes span representations using boundary and label queries over internal tokens, while triaffine scoring jointly models interactions to classify spans.

    \item Multiple LSTM-CRF (MLC) model~\cite{rojas-etal-2022-simple}: Trains separate sequence labeling models for each entity type independently, that simplifies nested NER.

    \item Multi-head Dense-augmented CRF~\cite{multi-head-crf}: Employs multi-head Dense-augmented CRF to learn the nested entities within biomedical datasets.
\end{itemize}

\begin{table*}[t!]
\centering
\caption{Summary statistics for the GENIA and Chilean Waiting List datasets, calculated using the official benchmark splits reported in~\cite{rojas-etal-2022-simple}.}
\label{table3}
\begin{tabular}{l rrr | rrr}
\toprule
 & \multicolumn{3}{c}{\textbf{GENIA}} & \multicolumn{3}{c}{\textbf{Chilean Waiting List}} \\
 \cmidrule(lr){2-4} \cmidrule(l){5-7}
\textbf{Metric} & \textbf{Train} & \textbf{Test} & \textbf{Val} & \textbf{Train} & \textbf{Test} & \textbf{Val} \\
\midrule
Sentences      & 15,023  & 1,854  & 1,669  & 8,014   & 990    & 890    \\
Tokens         & 454,882 & 57,021 & 48,932 & 149,574 & 18,436 & 16,754 \\
Entities       & 45,929  & 5,474  & 4,337  & 35,480  & 4,289  & 3,971  \\
Nested (\%)    & 17.0    & 20.6   & 16.8   & 46.4    & 45.9   & 46.7   \\
\bottomrule
\end{tabular}
\end{table*}

\subsection{Cross-Dataset Evaluation}
To assess the generalizability of our method, we further evaluate it on two widely-used NER datasets: GENIA and the Chilean Waiting List. This allows us to evaluate our approach across different domains. The statistics of these datasets and their train, test, and validation splits, as reported in prior work~\cite{rojas-etal-2022-simple}, are summarized in Table~\ref{table3}.

\paragraph{GENIA.}
The GENIA corpus is a widely used biomedical text dataset containing abstracts from PubMed related to transcription factors in human blood cells. It is annotated with nested named entities like proteins, DNAs, RNAs, cells, and cell lines. Due to its rich, overlapping annotations, GENIA is often used for evaluating NNER models in the biomedical domain.

\paragraph{Chilean Waiting List.}
The Chilean Waiting List dataset comprises clinical case descriptions (in Spanish) explaining the reasons for medical referrals within the Chilean public healthcare system. It features nested and complex medical entities related to diseases, symptoms, and procedures that makes it suitable for studying NNER in multilingual, real-world healthcare contexts.

\subsection{Hyperparamters Tuning}
We first train BERT–CRF models on the plasma physics dataset using manually selected hyperparameters commonly used in transformer-based NER, including a learning rate in the range $[1\times10^{-5},4\times10^{-5}]$, a dropout of $0.1$, a batch size in $[2, 32]$, a weight decay in $[0.01,0.03]$, and 15 training epochs. We then apply \textit{Bayesian Optimization} to tune the learning rate in $[10^{-6},10^{-4}]$, batch size in $[2, 32]$, and weight decay in $[0, 0.3]$, with the objective of maximizing the validation F1 score, and use the best configuration to train the final models.

\section{Results}
We report micro-averaged Precision, Recall, and F1 as the primary evaluation metrics, as they reflect overall extraction performance across all entity classes. In particular, we focus on \textit{strict span-based} evaluation, where a predicted entity is counted as correct only if both its span boundaries and its label match the gold annotation exactly. This evaluation protocol provides a more realistic and challenging assessment of nested named entity recognition performance.

\begin{table*}[t!]
\caption{Performance comparison of baseline models and the proposed method.}
\centering
\begin{tabular}{l c c c}
\hline
\textbf{Model} & \textbf{Precision} & \textbf{Recall} & \textbf{F1} \\ \hline
Layered BiLSTM-CRF \cite{ju-etal-2018-neural} & 0.63 & 0.54 & 0.58 \\
Multiple LSTM-CRF (MLC) \cite{rojas-etal-2022-simple} & 0.58 & 0.61 & 0.59 \\
Span-based and biaffine \cite{yan-etal-2023-embarrassingly} & 0.62 & 0.63 & 0.63 \\
Recursive-CRF \cite{shibuya2020nested} & 0.64 & 0.63 & 0.63 \\
Pyramid \cite{wang-etal-2020-pyramid} & 0.65 & 0.65 & 0.65 \\
Triaffine (Span-based Attention network) \cite{yuan-etal-2022-fusing} & 0.73 & 0.58 & 0.65 \\
Triaffine (Span-based) \cite{yuan-etal-2022-fusing} & 0.66 & 0.65 & 0.65 \\
Multi-head Dense-augmented CRF \cite{multi-head-crf} & 0.67 & 0.66 & 0.66 \\
Partially observed Tree CRFs \cite{Fu_Tan_2021} & \textbf{0.69} & 0.68 & \textbf{0.69} \\
\hline
Our method (BERT-CRF) - w/o BO & 0.61 & 0.71 & 0.65 \\
Our method - BO-optimized & 0.64 & \textbf{0.74} & 0.68 \\ \hline
\end{tabular}
\label{table4}
\end{table*}

Table~\ref{table4} presents a comparative evaluation of our proposed method against several state-of-the-art baselines on the plasma physics NNER dataset. Among the existing approaches, the Partially Observed Tree CRFs achieves the highest performance with an F1 score of 0.69, followed by Pyramid (0.65), Triaffine models (0.65), and Multi-head Dense-augmented CRF (0.66). These results show that span-based and tree-structured models remain competitive for nested entity recognition in this domain. Our BERT-CRF model without Bayesian Optimization achieves an F1 score of 0.65, which is comparable to Recursive-CRF, Span-based and Biaffine, and MLC baselines. When hyperparameters are optimized using Bayesian Optimization, our model achieves an F1 score of 0.68, comparable to the highest-performing baseline. While this does not exceed the strongest TreeCRF baseline, it demonstrates that a simpler architecture can achieve near state-of-the-art performance under strict span-based evaluation. Most importantly, our BO-optimized model achieves the highest recall among all compared methods (0.74). This indicates that our approach is particularly effective at identifying a larger proportion of relevant entities, which is especially desirable in knowledge graph construction. The results show that complex structured models provide only marginal gains despite higher computational costs, while BERT-CRF achieves competitive performance with a lightweight design.

{\setlength{\tabcolsep}{6pt}
\begin{table}[t!]
\centering
\caption{Performance comparison of the NNER model with manually selected hyperparameters vs. Bayesian Optimization (BO), reporting F1 per entity under two evaluation schemes: token-based and strict span-based.}
\resizebox{\textwidth}{!}{%
\begin{tabular}{l cc cc}
\hline
\multirow{2}{*}{\textbf{Entity Type}} 
& \multicolumn{2}{c}{\textbf{F1 (token-based)}} 
& \multicolumn{2}{c}{\textbf{F1 (span-based)}} \\
\cline{2-5}
& \textbf{w/o BO} & \textbf{with BO} & \textbf{w/o BO} & \textbf{with BO} \\
\hline
Physical Quantity        & 0.80 & 0.82 & 0.64 & 0.67 \\
Species                  & 0.84 & 0.84 & 0.68 & 0.70 \\
Diagnostic Device        & 0.79 & 0.82 & 0.63 & 0.66 \\
Power Supply             & 0.82 & 0.83 & 0.61 & 0.64 \\
Plasma Medium            & 0.80 & 0.84 & 0.62 & 0.68 \\
Experiment               & 0.80 & 0.85 & 0.65 & 0.69 \\
Electrode Material       & 0.82 & 0.84 & 0.66 & 0.68 \\
Plasma Application       & 0.75 & 0.80 & 0.58 & 0.60 \\
Unit                     & 0.94 & 0.94 & 0.87 & 0.87 \\
Physical Effect          & 0.81 & 0.83 & 0.60 & 0.63 \\
Modelling                & 0.75 & 0.81 & 0.55 & 0.60 \\
Discharge Regime         & 0.87 & 0.89 & 0.71 & 0.72 \\
Electrode Configuration  & 0.83 & 0.88 & 0.65 & 0.68 \\
Plasma Properties        & 0.77 & 0.87 & 0.68 & 0.75 \\
Plasma Target            & 0.46 & 0.70 & 0.44 & 0.64 \\
Plasma Source            & 0.84 & 0.88 & 0.75 & 0.78 \\
\hline
\end{tabular}%
}
\label{table5}
\end{table}
}

Table~\ref{table5} presents the results for each class, comparing models trained without and with Bayesian Optimization. The improvements demonstrate that automated hyperparameter tuning leads to better model performance. By applying Bayesian Optimization, we observe consistent improvements in F1 scores compared to models trained with fixed, manually selected hyperparameters. For most classes, the performance improved after tuning, while for a few, the optimized results remain same to the fixed setup. Notably, for under-performing classes such as \textit{Plasma Target}, Bayesian Optimization significantly enhanced the model's performance, increasing the token-based F1 score from 0.46 to 0.70 and the strict span-based F1 from 0.44 to 0.64. This demonstrates that systematically exploring the hyperparameter space produce more reliable configurations according to the unique characteristics of each class. As a result, training the model with optimized hyperparameters leads to better and more generalizable performance across diverse entity classes.

We also trained our model on publicly available datasets, namely GENIA and the Chilean Waiting List. Table~\ref{table6} compares the performance on these datasets. On GENIA, different baselines achieve F1-scores between 0.71 and 0.81, with Triaffine performing best, while our BERT-CRF attains competitive performance (F1 = 0.77). On the Chilean Waiting List, Pyramid, Recursive-CRF, and MLC report F1-scores in the range of 0.76–0.78, and our method achieves comparable results (F1 = 0.79). Overall, these results indicate that a lightweight BERT-CRF can remain competitive with more state-of-the-art architectures across standard NER benchmarks. Dashes indicate results not reported in the corresponding baseline studies.

\begin{table}[t!]
\centering
\caption{Performance comparison across GENIA and Chilean Waiting List datasets against baselines and our method.}
\begin{tabular}{l p{0.9cm}p{0.9cm}p{0.9cm} p{0.9cm}p{0.9cm}p{0.9cm}}
\hline
\multirow{2}{*}{\textbf{Model}} 
& \multicolumn{3}{c}{\textbf{GENIA}} 
& \multicolumn{3}{c}{\makebox[2.7cm][r]{\textbf{Chilean Waiting List}}} \\
\cline{2-7}
& \textbf{P} & \textbf{R} & \textbf{F1} 
& \textbf{P} & \textbf{R} & \textbf{F1} \\
\hline
Layered         & 0.73 & 0.68 & 0.71 & 0.75 & 0.72 & 0.73 \\
Exhaustive      & 0.74 & 0.69 & 0.71 & 0.76 & 0.71 & 0.68 \\
Pyramid         & 0.78 & 0.72 & 0.75 & 0.79 & 0.75 & 0.77 \\
Biaffine        & 0.79 & 0.73 & 0.76 & 0.81 & 0.67 & 0.73 \\
Recursive-CRF   & 0.75 & 0.75 & 0.75 & 0.75 & 0.77 & 0.76 \\
MLC             & 0.77 & 0.74 & 0.75 & 0.77 & 0.78 & 0.78 \\
PO-TreeCRFs     & 0.78 & 0.78 & 0.78 & - & - & - \\
Triaffine       & 0.80 & 0.82 & 0.81 & - & - & - \\
Span-level Graphs & 0.77 & 0.80 & 0.79 & - & - & - \\
Our method      & 0.77 & 0.77 & 0.77 & 0.79 & 0.78 & 0.79 \\
\hline
\end{tabular}
\label{table6}
\end{table}

\section{Conclusions}
\label{s:conclusion}
We propose a lightweight approach based on BERT-CRF to extract nested named entities from plasma physics articles. To support this task, we have annotated and published the plasma physics NNER dataset. This dataset enables the extraction of detailed information such as experiments information, diagnostic devices, plasma sources, and more. Our method adopts an entity-specific specialization setting, in which independent BERT–CRF models are trained for individual entity types and their predictions are aggregated to form the final output. In addition, we incorporate Bayesian Optimization to automatically fine-tune model hyperparameters to ensure the optimal performance of a trained model. Our results show that the proposed approach achieves competitive strict span-based performance on the plasma physics dataset while maintaining substantially lower architectural complexity than existing nested NER models. Our findings indicate that lightweight, well-tuned models can provide practical solutions for real-world scientific information extraction research problems.

\section{Data and Code Availability}
The scores of representative models are published in reproducible form in the TIB Knowledge Loom (KL) at \url{https://doi.org/10.82209/hv44-a941} following the KL method \cite{stocker25nsd}.

\section*{Acknowledgments}
The work was funded by the German Federal Ministry of Research, Technology and Space (BMFTR) under the grant marks 16KOA013A and 16KOA013B.

\section*{Disclosure of Interests}
The authors declare that they have no competing interests.

\bibliographystyle{splncs04}
\bibliography{references}
\end{document}